\documentclass{article}
\usepackage{graphicx}
\usepackage{ijcai25}

\usepackage{times}
\usepackage{soul}
\usepackage{url}
\usepackage[hidelinks]{hyperref}
\usepackage[utf8]{inputenc}
\usepackage[small]{caption}
\usepackage{graphicx}
\usepackage{amsmath}
\usepackage{amsthm}
\usepackage{booktabs}
\usepackage{algorithm}
\usepackage{algorithmic}
\usepackage[switch]{lineno}
\usepackage{amssymb}
\usepackage{multirow} 
\usepackage{xcolor}
\usepackage{colortbl}
\usepackage{hyperref}

\title{SpatiaLoc: Leveraging Multi-Level Spatial Enhanced Descriptors for Cross-Modal Localization}

\author{
Tianyi Shang$^1$, Pengjie Xu$^3$, Zhaojun Deng$^4$, Zhenyu Li$^{2}$\thanks{Corresponding author}, Zhicong Chen$^{1}$, Lijun Wu$^1$
\\
\affiliations
$^1$Fuzhou University\\
$^2$Shandong Academy of Sciences\\
$^3$Qingdao University\\
$^4$Tongji University
}
\begin{document}
\maketitle
\begin{abstract}
Cross-modal localization using text and point clouds enables robots to localize themselves via natural language descriptions, with applications in autonomous navigation and interaction between humans and robots. In this task, objects often recur across text and point clouds, making spatial relationships the most discriminative cues for localization. Given this characteristic, we present SpatiaLoc, a framework utilizing a coarse-to-fine strategy that emphasizes spatial relationships at both the instance and global levels. In the coarse stage, we introduce a Bezier Enhanced Object Spatial Encoder (BEOSE) that models spatial relationships at the instance level using quadratic Bezier curves. Additionally, a Frequency Aware Encoder (FAE) generates spatial representations in the frequency domain at the global level. In the fine stage, an Uncertainty Aware Gaussian Fine Localizer (UGFL) regresses 2D positions by modeling predictions as Gaussian distributions with a loss function aware of uncertainty. Extensive experiments on KITTI360Pose demonstrate that SpatiaLoc significantly outperforms existing state-of-the-art (SOTA) methods.
\end{abstract}
\section{Introduction}
Cross-modal localization using text and point clouds enables robots to perform localization using natural language descriptions, with applications in autonomous navigation, unmanned aerial vehicles, and delivery networks. Conventional visual visaul localization \cite{li2025place} relies on monocular images or LiDAR \cite{du20243} for database matching but faces two limitations: low efficiency in collaboration between humans and machines and degraded robustness under changes in time or viewpoint. Localization based on text and point clouds addresses these by directly correlating textual descriptions with point cloud spatial signatures, removing the need for physical proximity during localization. \\
\indent This task presents two challenges: (1) vague textual queries map ambiguously to multiple candidate regions in point cloud maps of large scale, and (2) descriptions of adjacent areas show minimal linguistic distinctions, complicating coordinate prediction with high precision. To address these, Text2Pos \cite{kolmet2022text2pos} proposed a pipeline progressing from coarse to fine stages: coarse localization divides maps into submaps for alignment, while fine localization fuses multimodal embeddings to regress coordinates within candidate submaps. RET \cite{wang2023text} introduced the Relation Enhanced Transformer to model relations at the instance level. Text2Loc \cite{xia2024text2loc} improved performance in the coarse stage via contrastive learning and reduced computation in the fine stage with a mechanism free of matching. \\
\indent Despite these advances, current methods still overlook a fundamental characteristic of this problem: in point cloud maps at a city scale with simple textual descriptions, identical object instances (e.g., poles, sidewalks, traffic lights, and signs) frequently recur across different locations. Crucially, such descriptions lack specific details about individual objects, making it unreliable to extract distinctive features akin to approaches used in traditional visual localization. \\
\indent We reexamine this problem and argue that relative spatial relationships between objects constitute the most discriminative features in this task. Furthermore, textual descriptions are rich in terms indicating spatial relations (e.g., ``The pose is south of a gray building''; ``The pose is on top of a dark green wall''), which explicitly encode geometric constraints. \\
\indent To address these issues, we propose SpatiaLoc, which follows the framework ranging from coarse to fine levels introduced by Text2Pos. In the coarse stage, SpatiaLoc extracts spatial relationships at both the instance and global levels, enabling more effective alignment across modalities between textual descriptions and point cloud submaps.This approach addresses a key limitation of prior methods, which were largely restricted to global-level alignment (i.e., generating only global descriptors) and consequently neglected crucial instance-level spatial relationships.  \\
\indent Specifically, in the coarse stage, we introduce a Bezier Enhanced Object Spatial Encoder (BEOSE) to refine explicit spatial relationships at the instance level via quadratic Bezier curves. This approach effectively mitigates the discrepancy where spatial features share directional similarity but exhibit distinct feature values during the alignment between text and point clouds. \\
\indent Concurrently, the Frequency Aware Encoder (FAE) projects raw point cloud feature sequences into the frequency domain to extract discriminative spatial geometric structures. Notably, even when employing the FAE in isolation (i.e., without instance-level features), our approach outperforms existing SOTA methods.\\
\indent In the fine stage, an Uncertainty Aware Gaussian Fine Localizer (UGFL) regresses 2D positions by modeling predictions as Gaussian distributions with a loss function aware of uncertainty, enabling robust localization under ambiguity across modalities. Furthermore, UGFL employs cascaded attention to recurrently update hidden states, thereby achieving effective cross-modal feature fusion. \\
\indent The contributions of our work is summarized as follows:
\begin{itemize}
    \item We propose SpatiaLoc, a novel coarse-to-fine framework that leverages dual-level spatial relationships (instance and global) as the most discriminative features for robust text-to-point cloud localization.
    \item In the coarse stage, we introduce the BEOSE to capture precise instance-level spatial relationships via quadratic Bezier curves.
    \item We propose a FAE to capture spatial geometric features within the frequency domain, generating discriminative global frequency representations.
    \item We design the UGFL in the fine stage. It achieves robust 2D localization using uncertainty-aware Gaussian modeling, incorporating recurrent hidden state updates for enhanced cross-modal feature fusion. 
\end{itemize}
\section{Related Work}
\subsection{Visual Localization}
Conventional 2D Visual localization methods typically employ aggregation strategies like the Vector of Locally Aggregated Descriptors (VLAD) \cite{arandjelovic2016netvlad} and Generalized Mean (GeM) to generate feature vectors from query images, with matching performed directly on these 2D features. To improve robustness, researchers have investigated the mutual relationship between feature descriptors and their assigned clusters. \cite{li20252} Concurrently, a substantial number of recent methods rely on end-to-end Transformer training for generating powerful global descriptors. \cite{wang2022transvpr,li2024feature} Furthermore, dedicated approaches have been introduced to specifically address VPR under challenging conditions, such as poor illumination \cite{li2024toward} and severe viewpoint changes \cite{berton2023eigenplaces}. More recently, some techniques have successfully extended the localization scope of VPR from the city-level to the continent-level. \cite{lindenberger2025scaling} Finally, robust position recognition can also be achieved through cross-modal query methods, including text to image \cite{shang2025bridging} or point cloud to image \cite{xu2024c2l} retrieval, enabling localization when the current visual image is unavailable.
\subsection{LiDAR Localization}
\indent LPR generates global descriptors from raw point clouds or BEV projections \cite{lin2024rsbev}. Early work like PointNetVLAD \cite{uy2018pointnetvlad} combined PointNet \cite{qi2017pointnet} and NetVLAD to aggregate point features into compact global representations. Other BEV approaches, including BevPlace \cite{luo2023bevplace} and I2P-Rec \cite{zheng2023i2p}, enhanced rotational invariance through geometric transformations and utilized multimodal projections to capture complementary spatial information. \\
\indent Other LPR methods operate directly on point cloud topology using attention graphs or 3D-CNNs. Examples include LPD-Net \cite{liu2019lpd} using graph structures to model local geometric patterns, DH3D \cite{du2020dh3d} and SOENet \cite{xia2021soe} for refined pose estimation, and InCloud \cite{knights2022incloud} for embedding stability in incremental learning scenarios. \\
\indent A subset of LPR systems leverages transformer architectures to capture long-range dependencies. TransLoc3D \cite{xu2021transloc3d} captures broad context via self-attention mechanisms, while NDT-Transformer \cite{zhou2021ndt} and PPT-Net \cite{hui2021pyramid} focus on NDT-processed data and multi-scale feature extraction through hierarchical pyramid structures, respectively. \\
\indent Approaches using sparse 3D convolutions, such as MinkLoc3D \cite{komorowski2021minkloc3d}, often achieve strong performance by efficiently processing large-scale point clouds. Recent improvements include LCDNet \cite{cattaneo2022lcdnet}, which employs optimal transport theory to enhance feature matching, and LoGG3D-Net \cite{vidanapathirana2022logg3d}, which introduces local consistency loss to improve descriptor discriminability across different viewpoints.
\subsection{Text to Point Cloud localization}
\indent Text-to-point cloud localization was pioneered by Text2Pos \cite{kolmet2022text2pos}, which proposed a coarse-to-fine localization pipeline. Subsequent works like RET \cite{wang2023text} (attention mechanisms) and Text2Loc \cite{xia2024text2loc} (contrastive learning, matching-free fine stage) improved performance. More recently, MambaPlace \cite{shang2024mambaplace} integrated state space models (SSMs). GOTPR \cite{jung2025gotpr} employ intermediate graph structures for efficiency but remains limited to coarse-stage localization. \\
\indent Recent efforts tackle real-world challenges like data dependency and cross-modal ambiguity. IFRP-T2P \cite{wang2024instance} achieves instance-free localization using an instance query extractor and RowColRPA/RPCA for relative position awareness. Uncertainty modeling is now central: PMSH \cite{feng2025partially} handles partially matching submaps by incorporating uncertainty and propagating it to coarse retrieval via an uncertainty-aware metric. Similarly, CMMLoc \cite{xu2025cmmloc} addresses partial text via an uncertainty-aware Cauchy-Mixture-Model (CMM) framework, enhanced by spatial consolidation and cardinal direction integration.
\begin{figure*}
    \centering
    \includegraphics[width=0.95\linewidth]{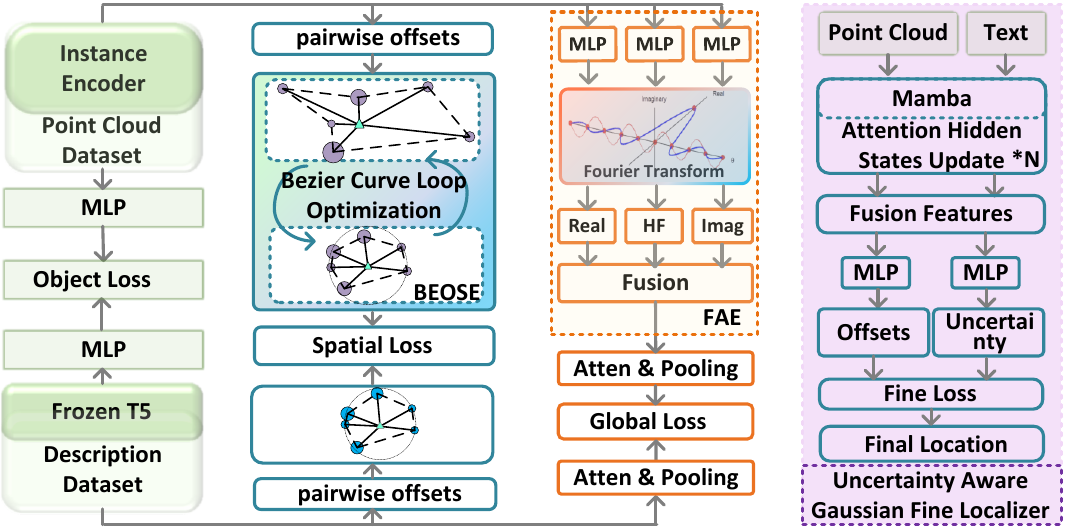}
    \vspace{-10pt}
    \caption{The overall architecture of the proposed SpatiaLoc. The left panel illustrates the coarse stage, which utilizes the BEOSE for instance-level spatial alignment and the FAE to extract frequency-domain spatial geometric features for global-level alignment. The right panel depicts the Fine Stage, employing the UGFL for precise position regression.}
    \label{Fig1}
\end{figure*}
\section{Methodology}
\subsection{Task Formulation}
Let $\mathcal{G} = \{R_i\}_{i=1}^{N_c}$ represent the gallery of point cloud submaps, where each submap $R_i = \{s_k\}_{k=1}^{N_s}$ encompasses $N_s$ object instances. The primary aim of this work is to determine the precise spatial location associated with a natural language query ${Q}$. Each query ${Q}$ comprises a sequence of $N_q$ relational descriptions, denoted as $\{d_j\}_{j=1}^{N_q}$. \\
\indent As shown in Figure \ref{Fig1}, We tackle this task through a coarse-to-fine strategy. In the coarse stage, the most relevant point cloud submap is retrieved using the textual query. In the fine stage, we determine the exact position ${L}_{gt} \in \mathbb{R}^2$ within the retrieved submap. \\
\indent This entire process is optimized by minimizing the distance between the ground-truth position and the predicted position over the data distribution $\mathcal{D}_s$:
\begin{equation}
\min_{\Psi,\,\Phi} \ \mathbb{E}_{({L_{gt}},\,{Q}) \sim \mathcal{D}_s}
\bigl\| {L}_{gt} - \Phi\!\left({Q}, \tilde{R}\right) \bigr\|_2^2,
\label{eq:task_obj_main}
\end{equation}
where $\tilde{R}$ is the submap retrieved from the gallery $\mathcal{G}$:
\begin{equation}
\tilde{R} = \arg\min_{R \in \mathcal{G}} \Delta \bigl( \Psi({Q}), \Psi(R) \bigr).
\label{eq:task_obj_retrieval}
\end{equation}
\indent Equation \eqref{eq:task_obj_main} describes the fine stage, where the regressor $\Phi$ fuses the query with the retrieved map $\tilde{R}$ to predict the final coordinates. Equation \eqref{eq:task_obj_retrieval} describes the coarse stage: the retrieval function $\Psi$ embeds the query and each submap into a shared latent space, uses the Euclidean distance $\Delta(\cdot,\cdot)$ to compute similarity, and selects the nearest neighbor.
\subsection{Object Level Alignment}
Following the definitions in the Task Formulation, given a query $Q$ with descriptions $\{d_j\}_{j=1}^{N_q}$ and a submap ${R_i}$ with instances $\{s_k\}_{k=1}^{N_s}$, we first extract their features. The descriptions are encoded via a frozen T5 encoder to obtain language features $\mathbf{T} = \{\mathbf{t}_j\}_{j=1}^{N_q}$, and the point cloud instances are encoded via a frozen PointNet++ to obtain visual features $\mathbf{V} = \{\mathbf{v}_k\}_{k=1}^{N_s}$.
\subsubsection{Relative Spatial Graph Construction}
We aim to address the ambiguity of natural language and repetitive object appearances. Since relative positional relationships between objects serve as the most discriminative features, we explicitly model these spatial relationships. We construct a pairwise offset tensor $\mathbf{O} \in \mathbb{R}^{N_s \times N_s \times 3}$:
\begin{equation}
\mathbf{O}_{mn} = \mathrm{Coord}(s_m) - \mathrm{Coord}(s_n),
\end{equation}
where $m, n \in \{1, \dots, N_s\}$ denote the instance indices, and $\mathrm{Coord}(\cdot)$ returns the instance centroid. $\mathbf{O}_{mn}$ represents the geometric vector from instance $n$ to $m$.
We fuse this geometric prior with semantic features to generate an initial edge representation $\mathbf{E}_{mn}$. Specifically, we concatenate the visual features of the instance pair with their spatial offset:
\begin{equation}
\mathbf{E}_{mn} = \mathrm{MLP}_{\text{fuse}}\Bigl( \bigl[ (\mathbf{v}_m \,,\, \mathbf{v}_n )\,;\, \mathrm{MLP}_{\text{geo}}(\mathbf{O}_{mn}) \bigr] \Bigr).
\end{equation}
\indent A coherent procedure is applied to the textual descriptions to obtain textual edge features $\mathbf{E}^t_{mn}$, relying on implicit directional cues embedded in the language features.
\subsubsection{Bezier-Enhanced Object Spatial Encoder (BEOSE)}
A modality gap exists in $\mathbf{E}^t_{mn}$ and $\mathbf{E}_{mn}$: $\mathbf{O}_{mn}$ encodes precise metric distances, whereas text typically conveys qualitative directional cues. Large absolute distances in $\mathbf{O}_{mn}$ can dominate the embedding, hindering alignment. To mitigate this, BEOSE modulates the spatial features via Bézier curves, retaining directionality while suppressing extreme magnitudes.
For each edge $\mathbf{E}_{mn}$, we employ a gated LSTM to extract a hidden state $\mathbf{h}_{mn} = \mathrm{LSTM}(\mathbf{E}_{mn})$. This state predicts three control points for a quadratic Bézier curve: two endpoints $\mathbf{P}^{(0)}, \mathbf{P}^{(2)}$ and an intermediate control $\mathbf{P}^{(1)}$:
\begin{equation}
\mathbf{P}^{(0)}_{mn} = \phi_0(\mathbf{h}_{mn}), \quad \mathbf{P}^{(2)}_{mn} = \phi_2(\mathbf{h}_{mn}), \\ \quad \mathbf{P}^{(1)}_{mn} = \phi_c(\mathbf{P}^{(0)}_{mn}),
\end{equation}
where $\phi(\cdot)$ denotes projection networks. Let $\tau_{mn} \in [0, 1]$ be a learnable interpolation parameter. The modulated edge feature $\hat{\mathbf{E}}_{mn}$ is synthesized as:
\begin{align}
\hat{\mathbf{E}}_{mn}(\tau_{mn}) = & (1-\tau_{mn})^2 \mathbf{P}^{(0)}_{mn} + 2(1-\tau_{mn})\tau_{mn} \mathbf{P}^{(1)}_{mn} \\ \notag
                                 + & \tau_{mn}^2 \mathbf{P}^{(2)}_{mn}.
\end{align}
\indent We enforce bounded displacement via $\mathbf{E}^{\star}_{mn} = \tanh(\hat{\mathbf{E}}_{mn})$, concluding one curve optimization. We iterate the above BEOSE steps $\mathcal{N}$ times. \\ 
\indent Figure \ref{Fig1} illustrates the modality gap between the encoded $\mathbf{E}^t_{mn}$ and $\mathbf{E}_{mn}$, along with the optimization process via BEOSE. As depicted, BEOSE is designed to suppress extreme magnitude values in $\mathbf{E}_{mn}$ while preserving directional characteristics, thereby mechanically aligning it with the textual features.
\subsubsection{Gaussian Aggregation (GA)}
To compress the pairwise edge features $\mathbf{E}^{\star} \in \mathbb{R}^{N_s \times N_s \times D}$ into node-level descriptors for alignment, we employ a Gaussian Aggregation module. We treat the features as probabilistic distributions. First, we generate a reparameterized latent variable $\mathbf{Z}_{mn}$:
\begin{equation}
\mathbf{Z}_{mn} = \boldsymbol{\mu}_{mn} + \exp(0.5 \log \boldsymbol{\sigma}^2_{mn}) \odot \boldsymbol{\epsilon}, \quad \boldsymbol{\epsilon} \sim \mathcal{N}(0, \mathbf{I}),
\end{equation}
where $\boldsymbol{\mu}$ and $\boldsymbol{\sigma}^2$ are predicted from $\mathbf{E}^{\star}_{mn}$. We then aggregate the edge information for each subject instance $m$ by summing over all object instances $n$:
\begin{equation}
\hat{\mathbf{v}}_m = \sum_{n=1}^{N_s} \left( \mathrm{Softmax}_n(\mathbf{Z}_{mn}) + \mathrm{Softmax}_n(\mathbf{E}^{\star}_{mn}) \right).
\end{equation}
\indent This yields a compact spatial descriptor set $\hat{\mathcal{V}} = \{\hat{\mathbf{v}}_m\}_{m=1}^{N_s}$ for the point cloud submap. 
An identical aggregation is performed on textual edges $\mathbf{E}^t_{mn}$ to obtain $\hat{\mathcal{T}} = \{\hat{\mathbf{t}}_j\}_{j=1}^{N_q}$.
\begin{table*}
\centering
\caption{A comprehensive comparison between SpatiaLoc and existing SOTA methods on the KITTI360Pose dataset. (Best results are \textbf{bolded}, and second-best results are \underline{underlined}.) For SpatiaLoc (Global), we rely solely on global features during the coarse stage, without incorporating instance-level features or their corresponding losses.}
\label{table:compare_with_sota}
\resizebox{\linewidth}{!}{
\begin{tabular}{ccccccc}
\hline
\multirow{2}{*}{Methods} & \multicolumn{6}{c}{Localization Recall ($\epsilon < 5/10/15m$) $\uparrow$} \\ \cline{2-7} 
 & \multicolumn{3}{c}{Validation Set} & \multicolumn{3}{c}{Test Set} \\ \cline{2-7} 
 & k = 1 & k = 5 & k = 10 & k = 1 & k = 5 & k = 10 \\ \hline
NetVLAD (\textcolor{blue}{CVPR'15}) & 0.18/0.33/0.43 & 0.29/0.50/0.61 & 0.34/0.59/0.69 & --- & --- & --- \\
PointNetVLAD (\textcolor{blue}{CVPR'18}) & 0.21/0.28/0.30 & 0.44/0.58/0.61 & 0.54/0.71/0.74 & 0.13/0.17/0.18 & 0.28/0.37/0.39 & 0.28/0.37/0.39 \\
Text2Pos (\textcolor{blue}{CVPR'22}) & 0.14/0.25/0.31 & 0.36/0.55/0.61 & 0.48/0.68/0.74 & 0.13/0.20/0.30 & 0.33/0.42/0.49 & 0.43/0.61/0.65 \\
RET (\textcolor{blue}{AAAI'23}) & 0.19/0.30/0.37 & 0.44/0.62/0.67 & 0.52/0.72/0.78 & 0.16/0.25/0.29 & 0.35/0.51/0.56 & 0.46/0.65/0.71 \\
Text2Loc (\textcolor{blue}{CVPR'24}) & 0.37/0.57/0.63 & 0.68/0.85/0.87 & 0.77/0.91/0.93 & 0.33/0.48/0.52 & 0.60/0.75/0.78 & 0.70/0.84/0.86 \\
IFRP-T2P (\textcolor{blue}{ACM MM’24}) & 0.23/0.45/0.53 & 0.53/0.70/0.81 & 0.64/0.86/0.89 & 0.22/0.40/0.46 & 0.47/0.68/0.73 & 0.58/0.78/0.82 \\
MambaPlace (\textcolor{blue}{IROS’25}) & \underline{0.45}/0.62/0.68 & \underline{0.75}/\underline{0.89}/0.90 & \underline{0.83}/\underline{0.94}/\underline{0.95} & 0.38/0.52/0.55 & 0.66/0.79/0.81 & 0.76/0.87/0.89 \\
CMMLoc (\textcolor{blue}{CVPR'25}) & 0.44/0.62/0.68 & \underline{0.75}/0.88/0.90 & \underline{0.83}/0.93/\underline{0.95} & 0.39/0.53/0.56 & 0.67/0.80/0.82 & 0.77/0.87/0.89 \\
PMSH (\textcolor{blue}{ICCV'25}) & 0.42/0.62/0.68 & \underline{0.75}/\underline{0.89}/0.90 & \underline{0.83}/\underline{0.94}/\underline{0.95} & 0.39/0.55/\underline{0.59} & 0.68/0.81/0.83 & \underline{0.78}/0.89/0.90 \\ \hline
SpatiaLoc (Global) & 0.44/\underline{0.63}/\underline{0.69} & 0.74/0.88/\underline{0.91} & 0.82/\underline{0.94}/\underline{0.95} & \underline{0.41}/\underline{0.56}/\underline{0.59} & \underline{0.69}/\underline{0.82}/\underline{0.84} & 0.77/\underline{0.90}/\underline{0.91} \\
\textbf{SpatiaLoc (coarse-to-fine)} & \textbf{0.54/0.77/0.82} & \textbf{0.81/0.95/0.97} & \textbf{0.86/0.98/0.98} & \textbf{0.51/0.71/0.74} & \textbf{0.78/0.91/0.92} & \textbf{0.83/0.95/0.96} \\ \hline
\end{tabular}
}
\end{table*}
\subsubsection{Cross-Modal Alignment Loss}
We align the features at both the Spatial Level (using aggregated relations $\hat{\mathcal{V}}, \hat{\mathcal{L}}$) and the Instance Level (using raw features $\mathbf{V}, \mathbf{L}$).
Consider a batch of size $B$. Let $\mathcal{X}_i$ and $\mathcal{Y}_j$ be the feature sets for point cloud submaps and descriptions. We define a bidirectional set-to-set similarity:
\begin{equation}
\mathcal{S}(\mathcal{X}_b, \mathcal{Y}_b) = \frac{\exp(\lambda_{ij}^{\mathcal{X}2 \mathcal{Y}}/\gamma)}{\sum_{k=1}^B \exp(\lambda_{ij}^{\mathcal{X}2 \mathcal{Y}}/\gamma)} + \frac{\exp(\lambda_{ij}^{\mathcal{Y}2 \mathcal{X}}/\gamma)}{\sum_{k=1}^B \exp(\lambda_{ij}^{\mathcal{Y}2 \mathcal{X}}/\gamma)},
\label{eq:relation_similarity}
\end{equation}
where $\gamma$ is a temperature parameter, $\lambda_{ij}^{\mathcal{X}2 \mathcal{Y}}$ represents the average of the maximum similarity scores between $\mathcal{X}_i$ and $\mathcal{Y}_j$, and $\lambda_{ij}^{\mathcal{Y}2 \mathcal{X}}$ represents the average of the maximum similarity scores between $\mathcal{Y}_i$ and $\mathcal{X}_i$. The alignment loss is:
\begin{equation}
\mathcal{L}_{align}(\mathcal{X}, \mathcal{Y}) = -\frac{1}{B} \sum_{b=1}^B \log \bigl( 1 - \mathcal{S}(\mathcal{X}_b, \mathcal{Y}_b) \bigr).
\label{eq:relation_loss}
\end{equation}
\indent The total instance level loss combines the spatial loss $\mathcal{L}_{IS} = \mathcal{L}_{align}(\hat{\mathcal{V}}, \hat{\mathcal{T}})$ and the instance loss $\mathcal{L}_{IO} = \mathcal{L}_{align}(\mathbf{V}', \mathbf{T}')$, where $\mathbf{V}', \mathbf{T}'$ are projections of the raw features.
\subsection{Global Level Alignment}
\indent In the coarse retrieval stage, we aim to align global descriptors from the city-scale point cloud gallery $\mathcal{G}_{\text{all}} = \{\mathbf{r}_i\}_{i=1}^{N_c}$ and the textual queries $\mathcal{Q}_{\text{all}} = \{{q}_i\}_{i=1}^K$. To bypass spatial misalignment issues, we introduce a Frequency Aware Encoder (FAE) that processes submap features $\mathbf{r}_i$ in the frequency domain. So that SpatiaLoc can more easily learn the spatial positional relationships from the frequency domain features, and ultimately obtain distinctive feature descriptors.
\indent The FAE first projects the submap features $\mathbf{r}_i$ into three latent subspaces via separate fully-connected layers, obtaining $\boldsymbol{\xi}^{(k)}_i$ ($k=1,2,3$). These are converted to the frequency domain via Fast Fourier Transform (FFT):
\begin{equation}
\mathcal{Z}^{(k)}_i[m] = \sum_{t=0}^{T-1} \boldsymbol{\xi}^{(k)}_i[t] \cdot e^{-j \frac{2\pi}{T} mt}, \quad k = 1,2,3.
\label{eq:fft_new}
\end{equation}
where $T$ represents the sequence length and $j$ denotes the imaginary unit.
\indent We then extract complementary features from these three branches: \\
\indent Real Branch: We extract the real part of the frequency domain after performing the Fourier transform: $\boldsymbol{\nu}^{(1)}_i = \left[ \Re( \mathcal{Z}^{(1)}_i ) \right]$.  $\Re(\cdot)$ represents the real part. \\
\indent Imaginary Branch: We extract the imaginary part of the frequency domain after performing the Fourier transform: $\boldsymbol{\nu}^{(2)}_i = \left[ \Im( \mathcal{Z}^{(2)}_i ) \right]$. $\Im(\cdot)$ represents the imaginary part.  \\
\indent High-Frequency Branch: The third branch preserves the top 30\% high-frequency bins using a binary high-pass mask $\mathcal{H}_m$ to filter the signal ($\widehat{\mathcal{Z}^{(3)}_i} = \mathcal{H}_m \odot \mathcal{Z}^{(3)}_i$). The final feature is recovered via Inverse Fast Fourier Transform (IFFT):
\begin{equation}
\mathcal{H}_m = \mathbb{I}(m \geq 0.7T), \quad \boldsymbol{\nu}^{(3)}_i = \Re \left( \text{IFFT} \left( \widehat{\mathcal{Z}^{(3)}_i} \right) \right).
\label{eq:high_freq_branch_new}
\end{equation}
\indent These features are integrated via a Triple Cross-Attention mechanism. After enhancing each $\boldsymbol{\nu}^{(k)}_i$ via shared self-attention, we compute cross-attention weights $\boldsymbol{\Omega}_1$ (between $\boldsymbol{\nu}^{(3)}_i, \boldsymbol{\nu}^{(1)}_i$) and $\boldsymbol{\Omega}_2$ (between $\boldsymbol{\nu}^{(3)}_i, \boldsymbol{\nu}^{(2)}_i$). The final representation $\boldsymbol{\kappa}_i$ is derived using the combined weight $\boldsymbol{\Omega}_{comb} = \boldsymbol{\Omega}_1 \odot \boldsymbol{\Omega}_2$:
\begin{equation}
\boldsymbol{\kappa}_i = \text{Attention}{\left( \boldsymbol{\Omega}_{comb}, \boldsymbol{\nu}^{(3)}_i \right)}.
\label{eq:triple_attention_new}
\end{equation}
\indent Then, $\boldsymbol{\kappa}_i$ is further processed by stacked LSTMs to yield the global point descriptor $\mathbf{P}^{glo}_i$.
\indent For the textual query $q_i$, we employ a stacked architecture consisting of attention and maxpooling layers to generate the global descriptor $\mathbf{Q}^{glo}_i$.
\indent The coarse stage optimization combines cross-modal alignment with unimodal discrimination:
\begin{align}
\mathcal{L}_{Global} = & \mathcal{L}_{align}(\mathbf{Q}^{glo}_i, \mathbf{P}^{glo}_i) + \mathcal{L}_{align}(\mathbf{Q}^{glo}_i, \mathbf{Q}^{glo}_i) \\ \notag
                   + & \mathcal{L}_{align}(\mathbf{P}^{glo}_i, \mathbf{P}^{glo}_i),
\label{eq:global_level_loss_new}
\end{align}
where $\mathcal{L}_{align}(\mathbf{Q}^{glo}_i, \mathbf{Q}^{glo}_i)$ and $\mathcal{L}_{align}(\mathbf{P}^{glo}_i, \mathbf{P}^{glo}_i)$ are used to maintain the feature discriminability of unimodal descriptors in the frequency domain, while $\mathcal{L}_{align}(\mathbf{Q}^{glo}_i, \mathbf{P}^{glo}_i)$ is used for cross-modal alignment. \\
\indent The total loss in the coarse stage is:
\begin{equation}
\mathcal{L}_{Coarse} = \mathcal{L}_{Global} + \mathcal{L}_{IS} + \mathcal{L}_{IO}.
\end{equation}
\subsection{Fine Stage} 
\indent In the fine stage, we propose the Uncertainty Aware Gaussian Fine Localizer (UGFL) to quantify the degree of cross-modal ambiguity between the query text and the semantic submap. This allows the network to adaptively adjust its optimization strategy based on varying levels of uncertainty. \\
\indent We initiate the process with feature fusion in the fine stage. To implement this, we adopt an alternating structure of Mamba and Cross Attention. Specifically, Cross Attention injects complementary modal information into Mamba’s hidden space, enabling the model to establish effective and expressive cross-modal fusion. Subsequently, the resulting fused feature, denoted as $\mathbf{f}_u$, is passed through two lightweight MLP to regress the final position. \\
\indent Because the predicted coordinate remains uncertain relative to the ground truth guidance, we treat the prediction process as modeling a Gaussian random distribution. To avoid the computational complexity of full covariance estimation, we simplify the distribution parameters by directly predicting the 2D position offsets $\boldsymbol{\delta}$ alongside a positive scalar reliability weight $\lambda$. Accordingly, the final predicted 2D coordinate $L_{pr}$ is derived by adding the estimated offsets $\boldsymbol{\delta}$ to the geometric center of each point cloud submap. In this context, $\lambda$ corresponds to the inverse variance (precision) of the Gaussian distribution, acting as a confidence indicator.
Consequently, the optimization objective is formulated to balance the prediction error and the uncertainty regularization:
\begin{equation}\mathcal{L}{reg} = \lambda \left| L{pr} - L_{gt} \right|_1 + \lambda^{-1},\label{eq:uncertainty_loss}\end{equation}
where $\left\| \cdot \right\|_1$ denotes the $L_1$ distance. The first term weights the regression error by the reliability $\lambda$, allowing the network to adaptively reduce the penalty for ambiguous samples (where the implicit variance is high). The second term $\lambda^{-1}$ serves as a regularization constraint derived from the Gaussian normalization factor, preventing the trivial solution where $\lambda$ converges to zero.
\section{Experiment}
\subsection{Experiment Setup}
\indent We evaluate SpatiaLoc on the KITTI360Pose dataset, which covers 15.51 km$^2$ across nine urban zones. The dataset is partitioned into training (5 zones, 28,689 queries), validation (1 zone, 3,187 queries), and testing (3 zones, 11,505 queries) subsets. All experiments were conducted on a 128-core AMD EPYC CPU and a single NVIDIA RTX V100 (32 GB) GPU. During the coarse stage, we train for 20 epochs with a batch size of 64 and a learning rate of 0.0005, generating 256-dimensional feature vectors. Subsequently, the fine stage is trained for 100 epochs with a batch size of 32 and a learning rate of 0.0003. These settings align with standard benchmarks for KITTI360Pose.
\subsection{Comparison with SOTA Methods}
\subsubsection{Overall Performance}
As presented in Table \ref{table:compare_with_sota}, SpatiaLoc achieves a significant lead across all evaluation metrics. Most notably on the challenging test set, SpatiaLoc outperforms the previous SOTA methods by substantial margins of 12\%/16\%/15\% in Top-1 recall at 5m/10m/15m thresholds, respectively. These comprehensive improvements empirically validate our core hypothesis: spatial positional relationships are the most effective features for text to point cloud localization. Guided by this insight, SpatiaLoc demonstrates exceptional effectiveness and robustness in complex urban environments. 
\begin{table}
\centering
\caption{Text-to-point-cloud-submap retrieval accuracy of SpatiaLoc and SOTA methods in coarse stage on the KITTI360Pose dataset.}
\label{table:comapre_fine_stage}
\resizebox{\linewidth}{!}{
\begin{tabular}{ccccccc}
\hline
{}{}{Methods} & \multicolumn{6}{c}{Submap Retrieval Recall $\uparrow$} \\ \cline{2-7}
& \multicolumn{3}{c}{Validation Set} & \multicolumn{3}{c}{Test Set} \\ \cline{2-7}
& \multicolumn{1}{c}{k = 1} & \multicolumn{1}{c}{k = 3} & \multicolumn{1}{c}{k = 5} & \multicolumn{1}{c}{k = 1} & \multicolumn{1}{c}{k = 3} & k = 5 \\ \hline
Text2Pos & \multicolumn{1}{c}{0.14} & \multicolumn{1}{c}{0.28} & \multicolumn{1}{c}{0.37} & \multicolumn{1}{c}{0.12} & \multicolumn{1}{c}{0.25} & 0.33 \\
RET & \multicolumn{1}{c}{0.18} & \multicolumn{1}{c}{0.34} & \multicolumn{1}{c}{0.44} & \multicolumn{1}{c}{0.15} & \multicolumn{1}{c}{0.29} & 0.37 \\
Text2Loc & \multicolumn{1}{c}{0.31} & \multicolumn{1}{c}{0.54} & \multicolumn{1}{c}{0.64} & \multicolumn{1}{c}{0.28} & \multicolumn{1}{c}{0.49} & 0.58 \\
IFRP-T2P & \multicolumn{1}{c}{0.24} & \multicolumn{1}{c}{0.46} & \multicolumn{1}{c}{0.57} & \multicolumn{1}{c}{0.23} & \multicolumn{1}{c}{0.39} & 0.48 \\
MambaPlace & \multicolumn{1}{c}{0.35} & \multicolumn{1}{c}{0.61} & \multicolumn{1}{c}{0.72} & \multicolumn{1}{c}{0.31} & \multicolumn{1}{c}{0.53} & 0.62 \\
CMMLoc & \multicolumn{1}{c}{0.35} & \multicolumn{1}{c}{0.61} & \multicolumn{1}{c}{\underline{0.73}} & \multicolumn{1}{c}{0.32} & \multicolumn{1}{c}{0.53} & 0.63 \\
PMSH & \multicolumn{1}{c}{\underline{0.37}} & \multicolumn{1}{c}{\underline{0.63}} & \multicolumn{1}{c}{\underline{0.73}} & \multicolumn{1}{c}{\underline{0.34}} & \multicolumn{1}{c}{\underline{0.56}} & \underline{0.65} \\
SpatiaLoc & \multicolumn{1}{c}{\textbf{0.52}} & \multicolumn{1}{c}{\textbf{0.78}} & \multicolumn{1}{c}{\textbf{0.86}} & \multicolumn{1}{c}{\textbf{0.48}} & \multicolumn{1}{c}{\textbf{0.72}} & \textbf{0.80} \\ \hline
\end{tabular}}
\end{table}
\subsubsection{Coarse Stage Performance}
\indent We subsequently perform a more detailed comparison across the {coarse} and {fine} stages. Specifically, in the coarse stage, {SpatiaLoc significantly surpasses all SOTA methods in Submap Retrieval Recall}. As demonstrated in Table \ref{table:comapre_fine_stage}, on the challenging test set, SpatiaLoc achieves a recall of {0.48} at $k=1$, representing a substantial absolute improvement of {14\%} over the previous best method, PMSH (0.34). The performance advantage becomes even more pronounced as $k$ increases; at $k=5$, SpatiaLoc's recall reaches {0.80}, which is also significantly higher than PMSH's 0.65.
This outcome fully validates the effectiveness of our strategies in the coarse stage: {explicitly modeling spatial relationships at the instance level} and {learning global spatial enhancements via frequency-domain global descriptors}. The substantial overall performance improvement of SpatiaLoc is primarily realized in the coarse stage. \\
\indent Notably, even our SpatiaLoc (Global) variant outperforms previous methods on the test set. Specifically, it achieves recalls of 0.51 and 0.71 at k=1 within 5m and 10m, exceeding the previous SOTA PMSH by 2\% and 1\%, respectively. This strongly validates the effectiveness of utilizing our FAE to capture spatial geometric structures in the frequency domain for cross-modal alignment.
\subsubsection{Fine Stage Performance}
\indent In the fine stage, SpatiaLoc also maintains competitive and leading performance among all SOTA two-stage methods, as shown in Table \ref{table:fine_stage_performance}. On the challenging test set, SpatiaLoc achieves the best Top-1 Localization Recall ($k=1$) of {0.51} and the highest Top-10 Recall ($k=10$) of {0.83}. Although the performance gain is less dramatic than in the coarse stage, this result validates the effectiveness of our fine stage strategies: cross-modal implicit state update for robust feature fusion and uncertainty-aware position regression for precise localization. This consistent superiority confirms SpatiaLoc's comprehensive effectiveness across both submap retrieval and fine position estimation.
\begin{table}
\centering
\caption{Fine Stage accuracy of SpatiaLoc and SOTA two stage methods on the KITTI360Pose dataset.}
\label{table:fine_stage_performance}
\resizebox{\linewidth}{!}{
\begin{tabular}{ccccccc}
\hline
{}{}{Methods} & \multicolumn{6}{c}{Localization   Recall ($\epsilon < 5m$) ↑}                                                                                                                                              \\ \cline{2-7} 
                         & \multicolumn{3}{c}{Validation Set}                                                                                      & \multicolumn{3}{c}{Test Set}                                            \\ \cline{2-7} 
                         & \multicolumn{1}{c}{k = 1}         & \multicolumn{1}{c}{k = 5}         & \multicolumn{1}{c}{k = 10}        & \multicolumn{1}{c}{k = 1}         & \multicolumn{1}{c}{k = 5}         & k=10        \\ \hline
Text2Pos                 & \multicolumn{1}{c}{0.45}          & \multicolumn{1}{c}{0.73}          & \multicolumn{1}{c}{0.79}          & \multicolumn{1}{c}{0.38}          & \multicolumn{1}{c}{0.63}          & 0.71        \\
RET                      & \multicolumn{1}{c}{0.46}          & \multicolumn{1}{c}{0.74}          & \multicolumn{1}{c}{0.79}          & \multicolumn{1}{c}{0.39}          & \multicolumn{1}{c}{0.64}          & 0.69        \\
Text2Loc                 & \multicolumn{1}{c}{0.53}          & \multicolumn{1}{c}{0.87}          & \multicolumn{1}{c}{0.87}          & \multicolumn{1}{c}{0.47}          & \multicolumn{1}{c}{0.75}          & 0.81        \\
MambaPlace               & \multicolumn{1}{c}{0.53}          & \multicolumn{1}{c}{0.88}          & \multicolumn{1}{c}{0.86}          & \multicolumn{1}{c}{0.48}          & \multicolumn{1}{c}{0.76}          & 0.81        \\
CMMLoc                   & \multicolumn{1}{c}{0.51}          & \multicolumn{1}{c}{0.86}          & \multicolumn{1}{c}{0.85}          & \multicolumn{1}{c}{0.45}          & \multicolumn{1}{c}{0.75}          & 0.79        \\
PMSH                     & \multicolumn{1}{c}{\textbf{0.54}} & \multicolumn{1}{c}{\underline{0.80}} & \multicolumn{1}{c}{\textbf{0.87}} & \multicolumn{1}{c}{\underline{0.49}} & \multicolumn{1}{c}{\textbf{0.78}} & \underline{0.82} \\
SpatiaLoc                & \multicolumn{1}{c}{\textbf{0.54}} & \multicolumn{1}{c}{\textbf{0.81}} & \multicolumn{1}{c}{\underline{0.86}} & \multicolumn{1}{c}{\textbf{0.51}} & \multicolumn{1}{c}{\textbf{0.78}} & \textbf{0.83} \\ \hline
\end{tabular}}
\end{table}
\begin{table}
\centering
\caption{Ablation Study For Coarse Stage Loss Function On Test Set.}
\label{table:ablation_coarse_loss}
\resizebox{0.75\linewidth}{!}{ 
\begin{tabular}{ccc|ccc} %
\hline
\multicolumn{3}{c|}{Methods} & \multicolumn{3}{c}{Retrieval Recall $\uparrow$} \\ \cline{1-6} 
$\mathcal{L}_{IS}$ & $\mathcal{L}_{IO}$ & $\mathcal{L}_{Global}$ & k=1 & k=3 & k=5 \\ \hline
\textbf{\checkmark} & & & 0.30 & 0.49 & 0.61 \\ 
& \textbf{\checkmark} & & 0.29 & 0.48 & 0.60 \\
& & \textbf{\checkmark} & 0.35 & 0.57 & 0.67 \\ 
\textbf{\checkmark} & \textbf{\checkmark} & & 0.32 & 0.52 & 0.65 \\ 
\textbf{\checkmark} & & \textbf{\checkmark} & 0.44 & 0.68 & 0.78 \\ 
& \textbf{\checkmark} & \textbf{\checkmark} & 0.43 & 0.66 & 0.75 \\ 
\textbf{\checkmark} & \textbf{\checkmark} & \textbf{\checkmark} & \textbf{0.48} & \textbf{0.72} & \textbf{0.80} \\ \hline
\end{tabular}}
\end{table}
\subsection{Ablation Study}
We conducted extensive ablation studies on SpatiaLoc, thereby demonstrating the effectiveness of our proposed architecture. 
\subsubsection{Ablation for Loss Function in Coarse Stage}
\indent We primarily conducted ablation studies on the coarse stage of SpatiaLoc. Specifically, we first performed a comprehensive analysis of the loss functions within this stage. \\
\indent As shown in Table \ref{table:ablation_coarse_loss}, we conducted a comprehensive ablation study on the three loss functions ($\mathcal{L}_{IS}$, $\mathcal{L}_{IO}$, and $\mathcal{L}_{Global}$), evaluating all seven possible combinations. \\
\indent Notably, even when exclusively utilizing $\mathcal{L}_{Global}$, where the model functions similarly to cross-modal retrieval methods relying solely on global descriptors, SpatiaLoc achieves recall rates of 0.35/0.57/0.67 at $k=1, 3, 5$, respectively. This performance with a single loss term already surpasses the previous SOTA method, PMSH (0.34/0.56/0.65). This strongly validates the effectiveness of our proposed FAE module, which extracts spatial relationships from point clouds in the frequency domain to generate high-quality global feature descriptors. \\
\indent Furthermore, we observe that the performance of using solely $\mathcal{L}_{IS}$ is higher than that of $\mathcal{L}_{IO}$, suggesting that instance-level spatial features possess strong discriminative power as cross-modal characteristics. However, when either $\mathcal{L}_{IS}$ or $\mathcal{L}_{IO}$ is combined with c, the performance improves substantially compared to utilizing $\mathcal{L}_{Global}$ in isolation. This significant gain demonstrates that global features and instance-level features are mutually complementary, working synergistically to enhance the overall retrieval accuracy. \\
\indent Finally, the optimal performance is achieved when all three loss functions are employed simultaneously. By jointly integrating global context with instance-level spatial relationships and object features, SpatiaLoc achieves a substantial performance improvement (reaching 0.48/0.72/0.80), fully validating the efficacy of our multi-level loss in coarse stage. \\
\indent Furthermore, to investigate the internal mechanism of $\mathcal{L}_{Global}$, we conducted further experiments. When retaining only $\mathcal{L}_{align}(\mathbf{P}^{glo}_i, \mathbf{P}^{glo}_i)$, the performance on the test set drops to 0.47/0.71/0.78. Compared to the complete $\mathcal{L}_{Global}$, these metrics represent decreases of 1\%, 1\%, and 2\%, respectively. These results strongly validate the effectiveness of the single-modal self-contrastive loss design within $\mathcal{L}_{Global}$, demonstrating that increasing the distance between unmatched features significantly enhances the discriminability of global features.
\subsubsection{Ablation for Key Components}
We performed a comprehensive ablation study to validate the effectiveness of the key components within SpatiaLoc and their contribution to the overall performance. \\
\indent First, the BEOSE proves critical to model performance. Experimental results in Table \ref{table:ablation_key_compents} indicate that removing this module causes a sharp 9\% decline in Recall@1 across both validation and test sets. This is attributed to BEOSE's ability to modulate spatial features via Bézier curves, which successfully suppresses the interference of absolute distance magnitudes on feature embedding while preserving directional semantics. \\
\indent Second, the absence of the FAE leads to significant performance degradation, with the test set R@1 dropping from 0.48 to 0.43. This confirms that the FAE effectively leverages frequency domain information to extract distinctive spatial positional features, thereby generating global descriptors that are more robust to environmental variations. \\
\indent Finally, removing the Gaussian Aggregation (GA) results in a 1\% decrease in test set at k=1. This demonstrates that, compared to conventional simple pooling operations, our Gaussian aggregation strategy combined with uncertainty sampling exhibits superior performance in feature dimensionality reduction and information compression.
\begin{table}
\centering
\caption{Ablation study of SpatiaLoc on the KITTI360Pose dataset. We individually remove the Bezier Enhanced Object Spatial Encoder (BEOSE), Frequency Aware Encoder (FAE), and Gaussian Aggregation (GA) to evaluate their contributions. w/o GA means that we replace GA with a simple maxpooling operation.}
\label{table:ablation_key_compents}
\resizebox{\linewidth}{!}{
\begin{tabular}{ccccccc}
\hline
{}{}{Methods} & \multicolumn{6}{c}{Submap Retrieval Recall $\uparrow$} \\ \cline{2-7}
& \multicolumn{3}{c}{Validation Set} & \multicolumn{3}{c}{Test Set} \\ \cline{2-7}
& \multicolumn{1}{c}{k = 1} & \multicolumn{1}{c}{k = 3} & \multicolumn{1}{c}{k = 5} & \multicolumn{1}{c}{k = 1} & \multicolumn{1}{c}{k = 3} & k = 5 \\ \hline
w/o BEOSE & \multicolumn{1}{c}{0.43} & \multicolumn{1}{c}{0.70} & \multicolumn{1}{c}{0.79} & \multicolumn{1}{c}{0.39} & \multicolumn{1}{c}{0.60} & 0.70 \\
w/o GA & \multicolumn{1}{c}{\underline{0.51}} & \multicolumn{1}{c}{\underline{0.77}} & \multicolumn{1}{c}{\underline{0.85}} & \multicolumn{1}{c}{\underline{0.47}} & \multicolumn{1}{c}{\underline{0.71}} & \underline{0.79} \\
w/o FAE & \multicolumn{1}{c}{0.48} & \multicolumn{1}{c}{0.74} & \multicolumn{1}{c}{0.83} & \multicolumn{1}{c}{0.43} & \multicolumn{1}{c}{0.65} & 0.74 \\
SpatiaLoc & \multicolumn{1}{c}{\textbf{0.52}} & \multicolumn{1}{c}{\textbf{0.78}} & \multicolumn{1}{c}{\textbf{0.86}} & \multicolumn{1}{c}{\textbf{0.48}} & \multicolumn{1}{c}{\textbf{0.72}} & \textbf{0.80} \\ \hline
\end{tabular}}
\end{table}
\subsubsection{Ablation for Fine Stage}
Furthermore, we conducted an ablation study targeting the fine stage. Specifically, after obtaining the fused features, we removed the MLP branch responsible for probability prediction. Consequently, in the absence of the probability term, the loss function degrades to solely computing the regression error between the predicted and ground-truth coordinates, which is consistent with the approach used in Text2Loc. Experimental results indicate that removing the probability module causes the performance metrics on the test set to drop from 0.51/0.78/0.83 to 0.48/0.77/0.81. This result strongly validates the effectiveness of the proposed UGFL in the fine stage.
\subsection{Visualization}
\begin{figure}
    \centering
    \includegraphics[width=\linewidth]{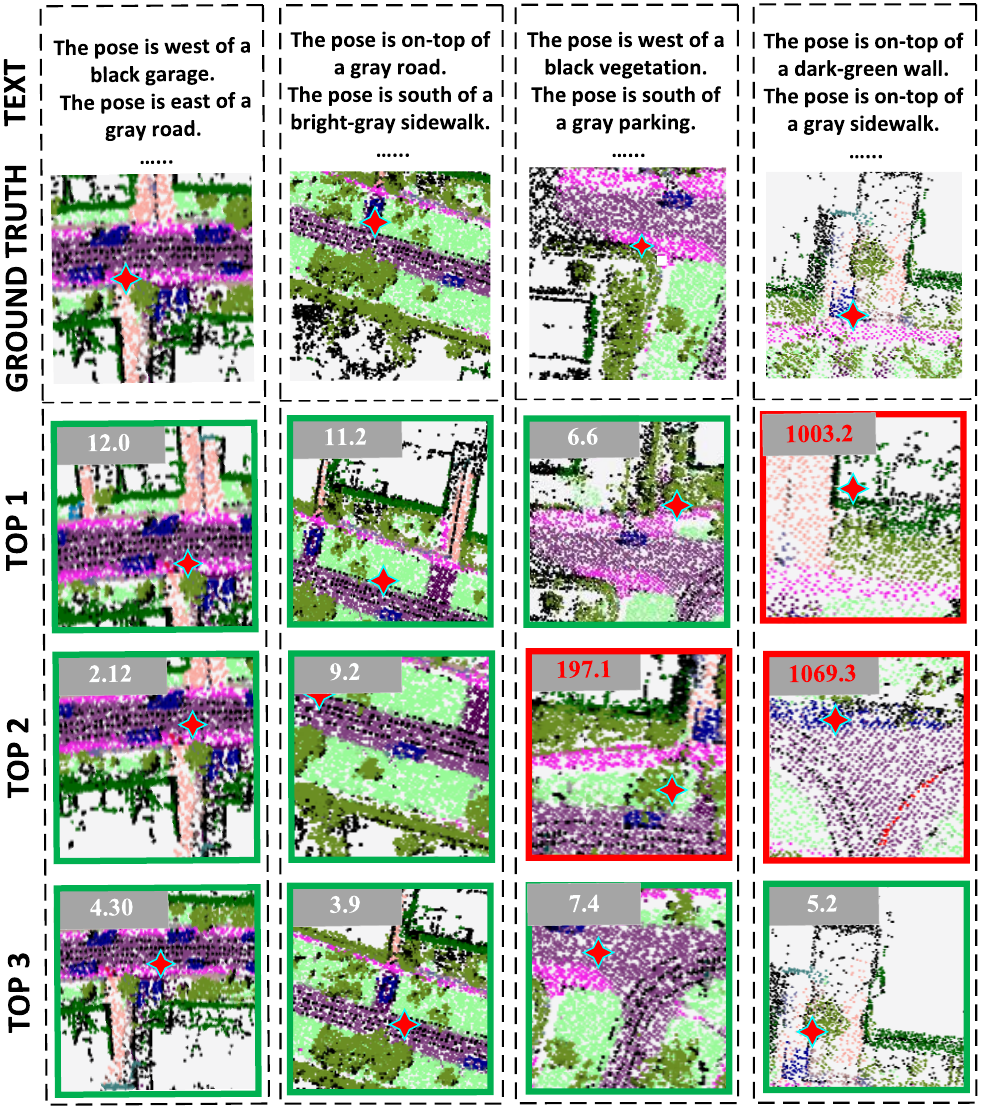}         
    \caption{Visualization Results for SpatiaLoc.}
    \label{Fig3}
\end{figure}

We visualize the localization results of SpatiaLoc in Figure \ref{Fig3}. We display the retrieved point cloud submaps and the specific coordinates regressed during the fine stage, indicated by red quadrangles. We also annotate the distance error between the final predicted position and the ground truth in the top-left corner of each result. The visualization results demonstrate that SpatiaLoc can effectively and accurately retrieve precise position coordinates in the vast majority of complex urban environments. However, due to viewpoint occlusion and the ambiguity of text descriptions, SpatiaLoc still encounters some errors.
\section{Conclusions}
In this paper, we presented SpatiaLoc, a novel coarse-to-fine framework for text-to-point cloud localization that identifies relative spatial relationships as the most discriminative cues for localization. To address the modality gap and object recurrence, we introduced the BEOSE to refine instance-level spatial relationships and FAE to extract robust global geometric features in the frequency domain. Furthermore, in the fine stage, we proposed the UGFL, which effectively handles cross-modal ambiguity through uncertainty-aware regression and recurrent feature fusion. Extensive experiments on the KITTI360Pose dataset demonstrate that SpatiaLoc significantly outperforms existing state-of-the-art methods in both submap retrieval and precise localization. In future work, we plan to enhance the model's robustness against severe occlusions and explore its application in larger-scale, unstructured environments.

\bibliographystyle{named}
\bibliography{ijcai25}

\end{document}